%% file: Template.tex
\documentclass{article}
\usepackage{amssymb}
\usepackage{spconf,amsmath,graphicx,hyperref}
\usepackage{multirow}
\usepackage{graphicx}
\usepackage{booktabs}
\title{Peeking Into The Future For Contextual Biasing}
\name{Ramaneswaran Selvakumar\sthanks{Work done during internship at Samsung Research America}, Cindy Tseng, Eesung Kim, Vijendra Raj Apsingekar, Yun Tang}
\address{Samsung Research America, USA}
\begin{document}
%\ninept
%
\maketitle
\begin{abstract}
While end-to-end (E2E) automatic speech recognition (ASR) models excel at general transcription, they struggle to recognize rare or unseen named entities (e.g., contact names, locations), which are critical for downstream applications like virtual assistants. In this paper, we propose a contextual biasing method for attention based encoder decoder (AED) models using a list of candidate named entities. Instead of predicting only the next token, we simultaneously predict multiple future tokens, enabling the model to "peek into the future" and score potential candidate entities in the entity list. Moreover, our approach leverages the multi-token prediction logits directly without requiring additional entity encoders or cross-attention layers, significantly reducing architectural complexity. Experiments on Librispeech demonstrate that our approach achieves up to 50.34\% relative improvement in named entity word error rate compared to the baseline AED model. 
\end{abstract}
\section{Introduction}
\label{sec:intro}

In recent years, end-to-end (E2E) automatic speech recognition (ASR) systems have emerged as the dominant paradigm for speech recognition~\cite{e2e_survey,li2022recentadvancesendtoendautomatic}. Unlike conventional systems that require individual components such as acoustic model (AM) and language model (LM), E2E approaches such as connectionist temporal classification (CTC)~\cite{graves_ctc,graves_asr} and attention encoder-decoder (AED)~\cite{radford2022robustspeechrecognitionlargescale} directly map acoustic features to text tokens. However, the effectiveness of these E2E approaches is inherently tied to the context present in its training data. Therefore, these models often fail to generalize to specific user contexts, as these contain information, particularly named entities that are rare or unseen during training. Named entities such as personal names or locations are so vast and diverse, that any static training dataset may not provide adequate coverage. Moreover, these named entities convey crucial semantic information to downstream natural language understanding (NLU) tasks in applications such as voice assistants. 

It is often impractical to retrain E2E-ASR models for every new context. Therefore, a common strategy is to utilize a 'bias list' - a list of keywords as external contextual knowledge to bias speech recognition. A simple approach to incorporate a bias list is shallow-fusion, which works by rescoring the ASR hypothesis during inference using an external biasing component such as another language model (LM)~\cite{huang2020classlmwordmapping, e2econtext_lm} or a weighted finite state transducer (WFST)~\cite{zhang2021tinytransducerhighlyefficientspeech, fox2022improvingcontextualrecognitionrare}. While this approach is architecturally simple, it requires a separate and often extensive tuning process to balance the scores from the main ASR model and the biasing component. Furthermore, despite this tuning effort, its contextualization capability is often limited, as the biasing is not deeply integrated into the model itself.

A more promising alternative is end-to-end neural biasing~\cite{pundak2018deepcontextendtoendcontextual,finecos}, which directly integrates the biasing component into the ASR model, enabling joint optimization. These methods use dedicated bias encoders to obtain bias phrase representations and implement cross-attention layers to associate bias phrases with the decoder or acoustic encoder. However, there are some limitations. It has been observed that simply adding a cross-attention layer for the bias list is not effective~\cite{huang2023contextualizedendtoendspeechrecognition}. To overcome this, some approaches use additional branches to detect bias phrases~\cite{huang2023contextualizedendtoendspeechrecognition, finecos}, deeper integration~\cite{chang2021contextawaretransformertransducerspeech, huang2024improvingneuralbiasingcontextual} or auxiliary loss functions~\cite{huber2021instantoneshotwordlearningcontextspecific, tang_improving_2024}, but this leads to more complex architectures which are difficult to deploy. Moreover, a common issue among these approaches is entity fragmentation~\cite{sudo_contextualized_2024}. They treat entities not as whole units, but as a disjointed sequence of sub-word tokens from the static vocabulary. For instance, a single entity like ``timaeus'' maybe fragmented into a sequence ``tim''~``ae''~``us'' forcing the model to learn the unnatural dependencies within the entity itself.

% However, it has been observed that simply adding a cross-attention layer for the bias list is not effective~\cite{huang2023contextualizedendtoendspeechrecognition}. To overcome this, some approaches use additional branches to detect bias phrases~\cite{huang2023contextualizedendtoendspeechrecognition, finecos} or auxiliary loss functions~\cite{huber2021instantoneshotwordlearningcontextspecific, tang_improving_2024}. While E2E neural biasing performs better than shallow-fusion, these architectures are more complex and difficult to deploy. Moreover, a common issue with these approaches is that during decoding they treat entities as a collection of individual tokens in the static vocabulary, rather than one complete unit. The consequence is that a whole entity such as ``Ronaldo'' maybe fragmented into a sequence ``Ron''~``al''~``do'' and the model is forced to learn the sub-word dependencies within this entity.

\noindent\textbf{Main Contributions:} In this paper, we propose an architecturally simple yet effective method of context biasing. First, we propose a multi-token prediction~\cite{gloeckle2024betterfasterlarge} (MTP) mechanism for the attention-encoder-decoder (AED) architecture. MTP enables the parallel prediction of multiple future tokens which allows the model to look-ahead and learn non-local sequence patterns. We then introduce a novel biasing mechanism that leverages this lookahead capability. Specifically, we utilize the un-normalized token logits and a learned projection to score entities in bias list. This approach does not require bias encoders or cross-attention layers, resulting in a simpler architecture. Finally, these entity scores are used as prediction logits, creating a unified search space at inference time, where at a time-step the model can either predict next token from its static vocabulary or next entity from the dynamic bias list. Moreover, by treating each entity as a single, indivisible candidate in the search space, our approach also mitigates the entity fragmentation problem. We demonstrate the effectiveness of our approach on Librispeech corpus where it reduces named entity WER by 50.34\% compared to the AED baseline.

\begin{figure}[htb]
  \centering
  \includegraphics[width=8.5cm]{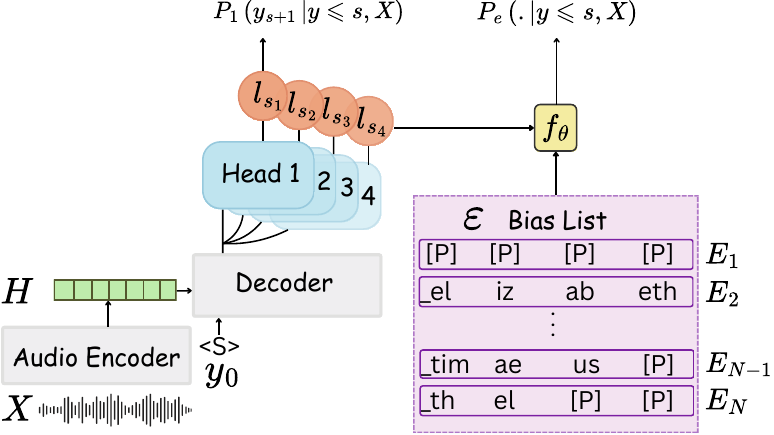}
  \caption{\small Overview of our proposed approach}
  \label{fig:biasing_weight}
\end{figure}

\section{Attention-Based Encoder Decoder}
\label{sec:aed_preliminary}

This section describes the attention-based encoder-decoder system that consists of an audio encoder and an attention-based decoder, which are extended to the proposed contextual biasing method.

\noindent\textbf{Audio Encoder} The audio encoder comprises two convolutional layers, a linear projection layer followed by several Conformer blocks [25]. The audio encoder transforms an audio feature sequence $X$ to $T$-length hidden state vectors $H_e = \text{AudioEnc}(X) = [h^e_1, ..., h^e_T] \in \mathbb{R}^{T \times d}$, where $d$ represents the dimension and $\text{AudioEnc}$ is the audio encoder.  %as follows:
%
%\begin{equation}
%H_e = \text{AudioEnc}(X).
%\end{equation}

\noindent\textbf{Attention-based Decoder:} The decoder is an autoregressive model that estimates the probability of the token sequence $y$ given the audio representation $H_e$. This is factorized as a product of conditional probabilities:

\begin{equation}
P(y | X) = \prod_{s=1}^S P(y_s | y_{0:s-1}, X).
\label{eq:ar_factorization}
\end{equation}

%where $s$ is the current time-step. 
At each decoding step $s$, the decoder network processes the previously generated tokens $y_{0:s-1}$ and the encoder output $H_e$ to produce a decoder output $h^d_s=\text{Dec}(y_{0:s-1}, H_e)$. 
The probability distribution for the next token $y_s$ is shown as below: %then obtained by applying a softmax function to these logits:

\begin{align}   
\mathbf{l}_{s} & = W_oh^d_s, \\
P(y_s | y_{0:s-1}, X) & = \text{Softmax}(\mathbf{l}_s),
\label{eq:decoder_output}
\end{align}
\noindent where $\mathbf{l}_s\in \mathbb{R}^V$ is output logits,  $V$ is the static vocabulary size, and $W_o$ is the final language model projection layer.

\section{Proposed Approach}

\subsection{Multi-Token Prediction}
\label{sec:multi_token_prediction}

The standard attention-based decoder from \autoref{sec:aed_preliminary} predicts tokens sequentially using $P(y_s | y_{0:s-1}, X) $, limiting its ability to consider future context when making current predictions. We propose to extend this model to predict $K$ future tokens simultaneously, enabling the model to "peek into the future" and evaluate long-term dependencies. This lookahead capability is particularly valuable for contextual biasing, as it allows better scoring of contextual entities by considering how well they fit within the broader sequence context.

To make the parallel prediction of a $K$-token sequence tractable, we employ a conditional independence assumption. The joint probability of the next $K$ tokens is approximated as the product of $K$ independent probabilities, all conditioned on the same preceding context:

\begin{align}
P(y_{s+1:s+K} | y_{\leq s}, X) &\approx \prod_{k=1}^K P_k(y_{s+k} | y_{\leq s}, X), \label{eq:mtp_approx}\\
\mathbf{l}^k_{s} & = g_k(h^d_s), \\
P(y_{s+k} | y_{0:s-1}, X) & = \text{Softmax}(\mathbf{l}^k_s),
\end{align}
\noindent where $g_{k}$ is the MTP module for the $k$-th prediction head, and $k \in [1,K]$.  $g_{k}$ consists of a Feed-Forward Network (FFN) block that generates a specialized representation for the $k^{th}$ future token.
 To ensure parameter efficiency, all heads share the final language model projection layer to produce the vocabulary logits.
Each $P_k(y_{s+k} | y_{\leq s}, X)$ can be computed in parallel during decoding.

\subsection{Entity Scoring}

At step $s$, the MTP mechanism from \autoref{sec:multi_token_prediction} produces $K$ future token logits simultaneously, yielding logit predictions $\mathbf{L}_s = [l^1_{s}, l^2_{s}, \dots, l^K_{s}] \in \mathbb{R}^{K \times V}$ where $V$ is the vocabulary size. We leverage these predictions to estimate the likelihood of each entity from a dynamic bias list $\mathcal{E}$ appearing as the next element in the sequence.

Given the bias list $\mathcal{E} = \{E_1, E_2, \dots, E_N\}$ of named entities where each entity $E_n$ is comprised of sub-words $E_n = [e_n^{(1)}, e_n^{(2)}, \dots, e_n^{(|E_n|)}]$. For each entity $E_n$ in the list, we construct an entity logit vector, $p_n$ by collecting the logit values from the model's future predictions that correspond to the entity's subword sequence. This is done by pairing the first subword of the entity, $e_n^{(1)}$, with the model's first future prediction, $l^1_{s}$; the second subword, $e_n^{(2)}$, with the second future prediction, $l^2_{s}$; and so on for the number of the MTP heads. If the number of tokens in an entity is less than $K$, we could use the padding token to fill the space; otherwise, if the number of tokens in an entity is more than $K$, we approximate the entity with its first $K$ tokens. The resulting vector is a direct alignment of the entity's sub-words with the model's future predictions

\begin{equation}
\mathbf{p}_n = \Bigl[l^1_{s}[e_n^{(1)}], l^2_{s}[e_n^{(2)}], \ldots, l^{K}_{s}[e_n^{(K)}]\Bigr] \in \mathbb{R}^{K}.
\end{equation}

$\mathcal{E}$ also contains a null entity $\emptyset$ comprised of padding tokens to denote absence of an entity. To get an entity likelihood score we use a trainable scoring function $f_\theta : \mathbb{R}^{K} \to \mathbb{R}$ which is realized using a FFN. If $z_{s}$ is the vector of entity scores ($z_{s} = f_{\theta}({p}_n)$) for all candidate entities in bias list at step $s$ then posterior probability of an entity being present at that step is:

\begin{equation}
P_{e}(\cdot | y{\leq s}, X) = \text{Softmax}(\mathbf{z}_s)
\label{eq:entity_prob}
\end{equation}

\subsection{Training}

The model is trained end-to-end by optimizing the sum of two loss functions: a multi-token prediction loss $\mathcal{L}_{\text{MTP}}$ and an entity scoring loss $\mathcal{L}_{\text{entity}}$. 

The objective for $\mathcal{L}_{\text{MTP}}$ is a multi-task cross-entropy loss that combines the predictions from all $K$ heads:

\begin{equation}
\mathcal{L}_{\text{MTP}} = - \frac{1}{S} \sum_{s=1}^S \sum_{k=1}^K \alpha_k \log P_k(y_{s+k} | y_{\leq s}, X)
\label{eq:mtp_loss_weighted}
\end{equation}

For the entity scoring loss $\mathcal{L}_{\text{entity}}$, we frame the task as a classification problem at each step $s$. To generate the supervision target, we define a ground truth label $\tau_s$  for each step. If a named entity $E_n$ located at index $n$ in bias list begins at step $s$ in the reference transcript, we set $\tau_s = n$ else its set to index of no-bias $\emptyset$, indicating that no entity starts at that step. 

\begin{equation}
\mathcal{L}_{\text{entity}} = - \frac{1}{S} \sum_{s=1}^S \log P_e(\tau_s | y_{\leq s}, X)
\label{eq:entity_loss}
\end{equation}

\subsection{Inference}

During inference, the model decodes from a unified search space that combines the static vocabulary and the dynamic bias list.  The unified search space $Q$ is formulated as:

\begin{equation}
Q(i) = \begin{cases}
P_e(\emptyset) \cdot P_{1}(i), & i \in V_{static} \\
\lambda \cdot P_e(i), & i \in \mathcal{E}
\end{cases}
\end{equation}

For brevity, we omit the conditioning on the $X$ and $y_{0:s-1}$ from our notation. Motivated by \cite{zhou2024copynebettercontextualasr} we use $P_e(\emptyset)$ as a prior probability that no entity is present at that step $s$ and use that to scale $P_{1}(i)$. Intuitively, this operation acts as a gating mechanism, if an entity gets a confident prediction, then $P_e(\emptyset)$ would be lower in value and suppress $P_{1}(i)$. We also use a manually tuned hyper-parameter $\lambda$ that acts as a biasing weight. We can increase this value to either boost context biasing or supress context biasing.

Furthermore, we apply a confidence threshold $\gamma$ to prune unlikely candidates during decoding. If the probability of the highest-scoring entity does not exceed this threshold ($\max_{E_n \in \mathcal{E}} P_e(E_n) < \gamma$) we set $P_e(E_n) = 0$ and $P_e(\emptyset) = 1$, effectively disabling context biasing and defaulting to just the static vocabulary

The final prediction uses greedy decoding over the unified search space:

\begin{equation}
\hat{y}_{s+1} = \arg\max_{i} Q(i | y_{0:s}, X).
\end{equation}

\section{Experimental Setup}

The input features are 80-dimensional log-Mel spectrograms with a window size of 400 samples and a hop length of 160 samples followed by SpecAugment. We use a conformer encoder with 12 layers, where each layer has hidden size of 512, expansion factor of 4 and 8 attention heads. The decoder contains 6 layers and similar to encoder, each layer has hidden size of 512, expansion factor of 4, 8 attention heads and pre-layer normalization. We use a pre-trained named entity recognition (NER) model\footnote{spaCy model: en\_core\_web\_trf} from spacy~\cite{ines_montani_2023_10009823} to annotate all named entities supported, except numerals\footnote{https://github.com/explosion/spaCy/discussions/9147}. There are total of 652/752 unique entities in clean/other sets. Across the two sets, the average number of tokens is 2.91 and 86.98\% of the entities have tokens less than or equal to 4. During training we randomly sample between 1-4 entities per sample to create an initial batch level bias list $B_{+}$ of size $|B_{+}|$. We further sample $\kappa \times |B_{+}|$ negative entities from the dataset to form a final bias list $B$ of size $N$. The proposed model trained for 100 epochs, with a learning rate of $1e^{-3}$ and 5500 warm-up steps using Adam~\cite{kingma2017adammethodstochasticoptimization} optimizer. After experimentations we finalize loss weight $\alpha_{k}$ in \autoref{eq:mtp_loss_weighted} as 1, 0.2, 0.1, 0.05 respectively and $\kappa$ as 2.

We implement our models in NeMo~\cite{nemo_toolkit} toolkit and our models are trained and evaluated on the Librispeech~\cite{librispeech} corpus. We re-implement and train CLAS~\cite{pundak2018deepcontextendtoendcontextual} with our AED model as baseline. Our primary evaluation metrics are word error rate (WER), biased WER (B-WER), and unbiased WER (U-WER)~\cite{le2021contextualizedstreamingendtoendspeech}. Here B-WER evaluates errors specifically within named entities and U-WER for the remaining words. The objective of our approach is to improve B-WER and prevent degradation of U-WER.  

\input{main_results}

\section{Results \& Discussion}

\begin{figure}[htb]
  \centering
  \includegraphics[width=6.5cm]{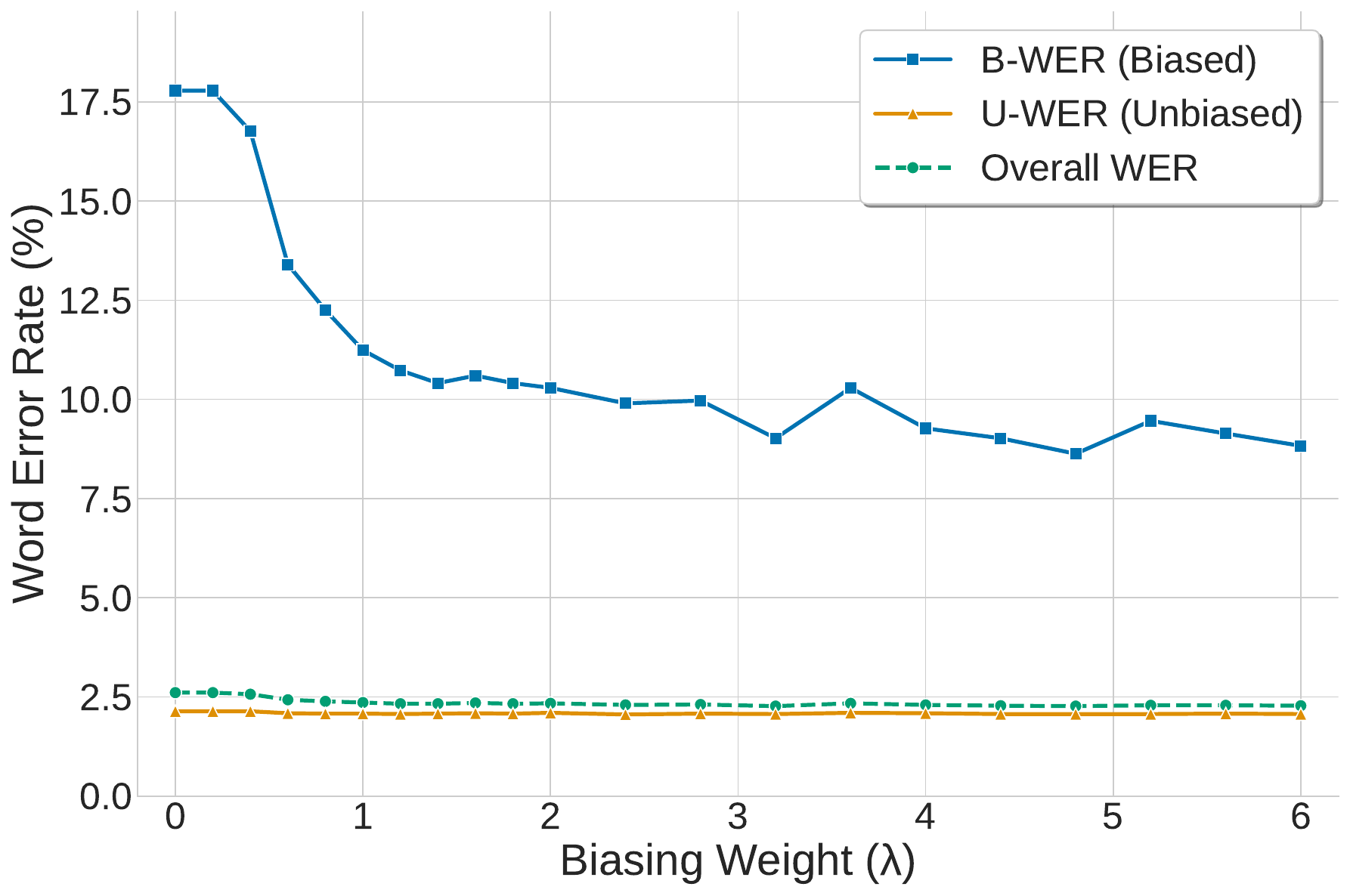}
  \caption{\small Effect of the biasing weight ($\lambda$) on word error rate on test-clean dataset. We can boost the context biasing performance to get further reduction in B-WER while U-WER remains stable.}
  \label{fig:biasing_weight}
\end{figure}

\noindent \textbf{Main Results} Tab~\ref{tab:main_results} shows the results obtained on the Librispeech corpus for different sizes of $B$. First, we observe that adding MTP to AED architecture either maintains or improves performance over baseline AED model. When $N = 100$, our proposed approach significantly improves the B-WER by upto 50.34\% in test-clean and 46.75\% in test-other. We notice that there is no significant degradation in WER when $N$ is scaled to 200 and 500. However, there is degradation in WER of test-other when no bias list is provided.

\noindent \textbf{Biasing Weight} \autoref{fig:biasing_weight} shows the effect of the biasing weight $\lambda$ which can boost or suppress the context biasing. Even with $\lambda$ of 1.0, our proposed approach significantly improves B-WER over our AED baseline, but further increasing $\lambda$ we see a further 30\% improvement in B-WER, while U-WER remains stable.

\noindent \textbf{MTP Head Architecture} Tab~\ref{tab:ablation_compiled} shows the WER when using a transformer based head for MTP ($g_k$) similar to \cite{gloeckle2024betterfasterlarge}. However, we observe that a FFN head (A1) performs betters than transformer head (A2), in fact transformer head shows degradation compared to AED (A0) baseline. 

\input{tab_ablation_compiled}

\noindent \textbf{Learned Entity Scoring} To validate the importance of the learned $f_{\theta}$, we replace it with a manual entity scoring function $f$, which is a weighted sum of logits ${p}_n$. Tab~\ref{tab:ablation_compiled} shows that $f$ (B1) can improve B-WER compared to AED baseline, indicating that MTP provides strong signals for selecting biasing entities. However, the results are not optimal and are worse than our approach (B0), moreover, it can also over-bias leading to poor overall WER compared to AED.

\noindent \textbf{Effect Of Number Of Heads} Next we check the importance of each head in driving the WER improvement. In Tab~\ref{tab:ablation_compiled} when we use only a single-head (B2) which is equivalent to using the regular next-token prediction for context biasing, we observe no improvement in B-WER compared to AED baseline, however when increasing the heads to 2 (B3) we observe some improvement in B-WER. This trend linearly increases with number of heads and in practice 4 heads provides a tradeoff between WER reduction and training complexity, given than in our case ~87\% of the entities are of less than 4 tokens.

\section{Conclusion}

In this paper we present a novel method for contextual biasing of speech recognition by leveraging multi-token prediction. We propose a dynamic vocabulary expansion method, where the raw logits of future token predictions are used to learn bias named entity scores. Unlike prior work which utilize bias encoders and cross-attention layers, our proposed method is simple yet effective. It reduces biased word error rate by up to 50.34\% on Librispeech corpus compared to the baseline attention encoder decoder model. We also investigate the impact of various components of our approach such as the architecture and number of multi-token heads, the importance of learned entity scoring and biasing weight that can be tuned to suppress or boost the context biasing.

% References should be produced using the bibtex program from suitable
% BiBTeX files (here: strings, refs, manuals). The IEEEbib.bst bibliography
% style file from IEEE produces unsorted bibliography list.
% -------------------------------------------------------------------------
\bibliographystyle{IEEEbib}
\bibliography{strings,refs}

\end{document}

%% file: main_results.tex
\begin{table*}[t]
    \centering
    
    \resizebox{\textwidth}{!}{% The resizebox command wraps the tabular environment
        \begin{tabular}{l|cc|cc|cc|cc}
\hline
\multicolumn{1}{c|}{\multirow{2}{*}{\textbf{Model}}}     & \multicolumn{2}{c|}{\textbf{N = 0 (no-bias)}}                                                                                          & \multicolumn{2}{c|}{\textbf{N = 100}}                                                                                                  & \multicolumn{2}{c|}{\textbf{N = 200}}                                                                                                  & \multicolumn{2}{c}{\textbf{N = 500}}                                                                                                   \\
\multicolumn{1}{c|}{}                                    & \textbf{test-clean}                                                          & \textbf{test-other}                                     & \textbf{test-clean}                                                          & \textbf{test-other}                                     & \textbf{test-clean}                                                          & \textbf{test-other}                                     & \textbf{test-clean}                                                          & \textbf{test-other}                                     \\ \hline
\begin{tabular}[c]{@{}l@{}}Baseline\\ (AED)\end{tabular} & \multicolumn{1}{c|}{\begin{tabular}[c]{@{}c@{}}2.73\\ (17.52/2.27)\end{tabular}} & \begin{tabular}[c]{@{}c@{}}6.01\\ (32.34/5.07
)\end{tabular} & \multicolumn{1}{c|}{\begin{tabular}[c]{@{}c@{}}-\end{tabular}} & \begin{tabular}[c]{@{}c@{}}-\end{tabular} & \multicolumn{1}{c|}{\begin{tabular}[c]{@{}c@{}}-\end{tabular}} & \begin{tabular}[c]{@{}c@{}}-\end{tabular} & \multicolumn{1}{c|}{\begin{tabular}[c]{@{}c@{}}-\end{tabular}} & \begin{tabular}[c]{@{}c@{}}-\end{tabular} \\ \hline
CLAS~\cite{pundak2018deepcontextendtoendcontextual}                                                    & \multicolumn{1}{c|}{\begin{tabular}[c]{@{}c@{}}3.12\\ (\textbf{15.56}/2.73)\end{tabular}} & \begin{tabular}[c]{@{}c@{}}6.95\\ (\textbf{28.26}/6.19)\end{tabular} & \multicolumn{1}{c|}{\begin{tabular}[c]{@{}c@{}}3.09\\ (15.56/2.70)\end{tabular}} & \begin{tabular}[c]{@{}c@{}}6.89\\ (27.81/6.14)\end{tabular} & \multicolumn{1}{c|}{\begin{tabular}[c]{@{}c@{}}3.08\\ (15.11/2.71)\end{tabular}} & \begin{tabular}[c]{@{}c@{}}6.89\\ (27.59/6.15)\end{tabular} & \multicolumn{1}{c|}{\begin{tabular}[c]{@{}c@{}}3.13\\ (15.75/2.74)\end{tabular}} & \begin{tabular}[c]{@{}c@{}}6.93\\ (27.92/6.18)\end{tabular} \\ \hline
AED-MTP                                                     & \multicolumn{1}{c|}{\begin{tabular}[c]{@{}c@{}}\textbf{2.58}\\ (17.27/2.27)\end{tabular}} & \begin{tabular}[c]{@{}c@{}}\textbf{6.00}\\ (30.63/5.12)\end{tabular} & \multicolumn{1}{c|}{\begin{tabular}[c]{@{}c@{}}2.58\\ (17.27/2.27)\end{tabular}} & \begin{tabular}[c]{@{}c@{}}6.00\\ (30.63/5.12)\end{tabular} & \multicolumn{1}{c|}{\begin{tabular}[c]{@{}c@{}}2.58\\ (17.27/2.27)\end{tabular}} & \begin{tabular}[c]{@{}c@{}}6.00\\ (30.63/5.12)\end{tabular} & \multicolumn{1}{c|}{\begin{tabular}[c]{@{}c@{}}2.58\\ (17.27/2.27)\end{tabular}} & \begin{tabular}[c]{@{}c@{}}6.00\\ (30.63/5.12)\end{tabular} \\ \hline
Ours($\lambda=1$)                                                     & \multicolumn{1}{c|}{\begin{tabular}[c]{@{}c@{}}2.61\\ (17.78/2.13)\end{tabular}} & \begin{tabular}[c]{@{}c@{}}6.25\\ (32.40/5.32)\end{tabular} & \multicolumn{1}{c|}{\begin{tabular}[c]{@{}c@{}}2.34\\ (10.98/2.07)\end{tabular}} & \begin{tabular}[c]{@{}c@{}}5.82\\ (21.85/5.24)\end{tabular} & \multicolumn{1}{c|}{\begin{tabular}[c]{@{}c@{}}2.36\\ (11.24/2.08)\end{tabular}} & \begin{tabular}[c]{@{}c@{}}5.82\\ (21.85/5.24)\end{tabular} & \multicolumn{1}{c|}{\begin{tabular}[c]{@{}c@{}}2.37\\ (11.49/2.09)\end{tabular}} & \begin{tabular}[c]{@{}c@{}}5.88\\ (22.68/5.27)\end{tabular} \\ \hline
Ours($\lambda=4.4$)                                                    & \multicolumn{1}{c|}{\begin{tabular}[c]{@{}c@{}}2.61\\ (17.78/2.13)\end{tabular}} & \begin{tabular}[c]{@{}c@{}}6.25\\ (32.40/5.32)\end{tabular} & \multicolumn{1}{c|}{\begin{tabular}[c]{@{}c@{}}\textbf{2.27}\\ (\textbf{8.70}/\textbf{2.07})\end{tabular}} & \begin{tabular}[c]{@{}c@{}}\textbf{5.64}\\ (\textbf{17.22}/5.22)\end{tabular} & \multicolumn{1}{c|}{\begin{tabular}[c]{@{}c@{}}\textbf{2.28}\\ (\textbf{9.02}/\textbf{2.07})\end{tabular}} & \begin{tabular}[c]{@{}c@{}}\textbf{5.63}\\ (\textbf{17.16}/5.22)\end{tabular} & \multicolumn{1}{c|}{\begin{tabular}[c]{@{}c@{}}\textbf{2.30}\\ (\textbf{9.27}/\textbf{2.08})\end{tabular}} & \begin{tabular}[c]{@{}c@{}}\textbf{5.64}\\ (\textbf{17.55}/5.21)\end{tabular} \\ \hline
\end{tabular}
    } % End of resizebox
\caption{\small WER (B-WER/U-WER) results on Librispeech-960 . \textbf{Bold} values represent the best result among the same bias list size N}\label{tab:main_results}
\end{table*}

%% file: tab_ablation_compiled.tex
\begin{table}[]
    \centering
    %\small
    %\resizebox{\columnwidth}{!}{% The resizebox command wraps the tabular environment
\begin{tabular}{@{}clcc@{}}
\toprule
\textbf{ID}            & \multicolumn{1}{c}{\textbf{Model}} & \textbf{test-clean} & \textbf{test-other} \\ \midrule
\multicolumn{1}{l}{A0} & AED                                & 2.73/17.52          & 6.01/32.34          \\ \midrule
A1                     & MTP (Linear)                       & 2.58/17.27          & 6.00/30.63          \\
A2                     & MTP (Transformer)                  & 2.99/19.49          & 6.98/37.03          \\ \midrule
B0                     & A0 + learned $f_{\theta}$                & 2.36/11.24          & 5.82/21.85          \\
B1                     & A0 + heuristic $f$              & 2.46/12.38          & 6.14/24.89          \\
B2                     & B0 + 1 MTP                         & 2.61/17.71          & 6.26/32.06          \\
B3                     & B0 + 2 MTP                         & 2.53/15.87          & 6.16/29.30          \\ \bottomrule
\end{tabular}
%}
\caption{\small Ablation on key architectural components (WER/B-WER)}\label{tab:ablation_compiled}
\end{table}

%% file: refs.bib
@inproceedings{zhou2024copynebettercontextualasr,
    title = "{C}opy{NE}: Better Contextual {ASR} by Copying Named Entities",
    author = "Zhou, Shilin  and
      Li, Zhenghua  and
      Hong, Yu  and
      Zhang, Min  and
      Wang, Zhefeng  and
      Huai, Baoxing",
    editor = "Ku, Lun-Wei  and
      Martins, Andre  and
      Srikumar, Vivek",
    booktitle = "Proc. ACL",
    month = aug,
    year = "2024",
    address = "Bangkok, Thailand",
    publisher = "Association for Computational Linguistics",
    url = "https://aclanthology.org/2024.acl-long.147/",
    doi = "10.18653/v1/2024.acl-long.147"
}

@inproceedings{sudo_contextualized_2024,
  title={Contextualized automatic speech recognition with dynamic vocabulary},
  author={Sudo, Yui and Fukumoto, Yosuke and Shakeel, Muhammad and Peng, Yifan and Watanabe, Shinji},
  booktitle={Proc. SLT},
  year={2024},
  organization={IEEE}
}

@article{huang2024improvingneuralbiasingcontextual,
  title={Improving neural biasing for contextual speech recognition by early context injection and text perturbation},
  author={Huang, Ruizhe and Yarmohammadi, Mahsa and Khudanpur, Sanjeev and Povey, Daniel},
  journal={arXiv preprint arXiv:2407.10303},
  year={2024}
}

@inproceedings{chang2021contextawaretransformertransducerspeech,
  title={Context-aware transformer transducer for speech recognition},
  author={Chang, Feng-Ju and Liu, Jing and Radfar, Martin and Mouchtaris, Athanasios and Omologo, Maurizio and Rastrow, Ariya and Kunzmann, Siegfried},
  booktitle={Proc. ASRU},
  year={2021},
  organization={IEEE}
}

@inproceedings{gloeckle2024betterfasterlarge,
author = {Gloeckle, Fabian and Idrissi, Badr Youbi and Rozi\`{e}re, Baptiste and Lopez-Paz, David and Synnaeve, Gabriel},
title = {Better \& faster large language models via multi-token prediction},
year = {2024},
publisher = {JMLR.org},
booktitle = {ICML},
articleno = {629},
numpages = {29},
location = {Vienna, Austria},
series = {ICML'24}
}

@inproceedings{le2021contextualizedstreamingendtoendspeech,
  author = {Le, Duc  and others},
  booktitle = {Interspeech},
  ee = {https://doi.org/10.21437/Interspeech.2021-1566},
  publisher = {ISCA},
  title = {Contextualized Streaming End-to-End Speech Recognition with Trie-Based Deep Biasing and Shallow Fusion.},
  url = {http://dblp.uni-trier.de/db/conf/interspeech/interspeech2021.html#LeJKKSMCSFKSS21},
  year = 2021
}

@INPROCEEDINGS{librispeech,
  author={Panayotov, Vassil and Chen, Guoguo and Povey, Daniel and Khudanpur, Sanjeev},
  booktitle={Proc. ICASSP}, 
  title={Librispeech: An ASR corpus based on public domain audio books}, 
  year={2015},
  volume={},
  number={},
  doi={10.1109/ICASSP.2015.7178964}}

@misc{nemo_toolkit,
  title        = {NeMo: a toolkit for Conversational AI and Large Language Models},
  author       = {Eric Harper et al.},
  howpublished = {\url{https://nvidia.github.io/NeMo/}},
  note         = {Code repository: \url{https://github.com/NVIDIA/NeMo}},
  year         = {2023}
}

@software{ines_montani_2023_10009823,
  author       = {Ines Montani and
                  Matthew Honnibal and
                  Matthew Honnibal and
                  Adriane Boyd and
                  Sofie Van Landeghem and
                  Henning Peters},
  title        = {explosion/spaCy: v3.7.2: Fixes for APIs and
                   requirements
                  },
  month        = oct,
  year         = 2023,
  publisher    = {Zenodo},
  version      = {v3.7.2},
  doi          = {10.5281/zenodo.10009823},
  url          = {https://doi.org/10.5281/zenodo.10009823},
}

@article{kingma2017adammethodstochasticoptimization,
  title={Adam: A Method for Stochastic Optimization},
  author={Diederik P. Kingma and Jimmy Ba},
  journal={CoRR},
  year={2014},
  volume={abs/1412.6980},
  url={https://api.semanticscholar.org/CorpusID:6628106}
}

@inproceedings{tang_improving_2024,
  title={Improving ASR contextual biasing with guided attention},
  author={Tang, Jiyang and Kim, Kwangyoun and Shon, Suwon and Wu, Felix and Sridhar, Prashant},
  booktitle={Proc. ICASSP},
  year={2024},
  organization={IEEE}
}

@inproceedings{huber2021instantoneshotwordlearningcontextspecific,
  title={Instant one-shot word-learning for context-specific neural sequence-to-sequence speech recognition},
  author={Huber, Christian and Hussain, Juan and St{\"u}ker, Sebastian and Waibel, Alexander},
  booktitle={Proc. ASRU},
  year={2021},
  organization={IEEE}
}

@article{huang2023contextualizedendtoendspeechrecognition,
  title={Contextualized end-to-end speech recognition with contextual phrase prediction network},
  author={Huang, Kaixun and Zhang, Ao and Yang, Zhanheng and Guo, Pengcheng and Mu, Bingshen and Xu, Tianyi and Xie, Lei},
  journal={arXiv preprint},
  year={2023}
}

@inproceedings{finecos,
  title={Improving end-to-end contextual speech recognition with fine-grained contextual knowledge selection},
  author={Han, Minglun and Dong, Linhao and Liang, Zhenlin and Cai, Meng and Zhou, Shiyu and Ma, Zejun and Xu, Bo},
  booktitle={Proc. ICASSP},
  year={2022},
  organization={IEEE}
}

@inproceedings{zhang2021tinytransducerhighlyefficientspeech,
  title={Tiny transducer: A highly-efficient speech recognition model on edge devices},
  author={Zhang, Yuekai and Sun, Sining and Ma, Long},
  booktitle={Proc. ICASSP},
  year={2021},
  organization={IEEE}
}

@article{fox2022improvingcontextualrecognitionrare,
  title={Improving contextual recognition of rare words with an alternate spelling prediction model},
  author={Fox, Jennifer Drexler and Delworth, Natalie},
  journal={arXiv preprint},
  year={2022}
}

@inproceedings{e2econtext_lm,
  title={Contextual Speech Recognition in End-to-end Neural Network Systems Using Beam Search.},
  author={Williams, Ian and Kannan, Anjuli and Aleksic, Petar S and Rybach, David and Sainath, Tara N},
  booktitle={Interspeech},
  year={2018}
}

@article{huang2020classlmwordmapping,
  title={Class LM and word mapping for contextual biasing in end-to-end ASR},
  author={Huang, Rongqing and Abdel-Hamid, Ossama and Li, Xinwei and Evermann, Gunnar},
  journal={arXiv preprint},
  year={2020}
}

@inproceedings{radford2022robustspeechrecognitionlargescale,
  title={Robust speech recognition via large-scale weak supervision},
  author={Radford, Alec and Kim, Jong Wook and Xu, Tao and Brockman, Greg and McLeavey, Christine and Sutskever, Ilya},
  booktitle={ICML},
  year={2023},
  organization={PMLR}
}

@inproceedings{graves_asr,
  title={Towards end-to-end speech recognition with recurrent neural networks},
  author={Graves, Alex and Jaitly, Navdeep},
  booktitle={ICML},
  year={2014},
  organization={PMLR}
}

@inproceedings{graves_ctc,
  title={Connectionist temporal classification: labelling unsegmented sequence data with recurrent neural networks},
  author={Graves, Alex and Fern{\'a}ndez, Santiago and Gomez, Faustino and Schmidhuber, J{\"u}rgen},
  booktitle={Proc. of 23rd ICML},
  year={2006}
}

@article{li2022recentadvancesendtoendautomatic,
  title={Recent advances in end-to-end automatic speech recognition},
  author={Li, Jinyu},
  journal={APSIPA Transactions on Signal and Information Processing},
  volume={11},
  number={1},
  year={2022},
  publisher={Now Publishers, Inc.}
}

@article{e2e_survey,
title = "End-to-End Speech Recognition: A Survey",
author = "Rohit Prabhavalkar and Takaaki Hori and Sainath, Tara N and Ralf Schluter and Shinji Watanabe",
note = "Publisher Copyright: {\textcopyright} 2014 IEEE.",
year = "2024",
doi = "10.1109/TASLP.2023.3328283",
language = "English",
volume = "32",
journal = "IEEE/ACM Transactions on Audio Speech and Language Processing",
issn = "2329-9290",
publisher = "IEEE Advancing Technology for Humanity",
}

@inproceedings{pundak2018deepcontextendtoendcontextual,
  title={Deep context: end-to-end contextual speech recognition},
  author={Pundak, Golan and Sainath, Tara N and Prabhavalkar, Rohit and Kannan, Anjuli and Zhao, Ding},
  booktitle={2018 Proc. IEEE SLT},
  year={2018},
  organization={IEEE}
}
